%% file: main.tex
\documentclass[nohyperref]{article}

\input{math_commands.tex}

\usepackage{microtype}
\usepackage{graphicx}
\usepackage{subfigure}
\usepackage{caption}
\usepackage{booktabs} 
\usepackage{makecell}
\usepackage{multicol}
\usepackage{multirow}
\usepackage{float}
\usepackage{enumitem}
\usepackage{wrapfig}
\usepackage{algorithmic}
\usepackage[ruled, linesnumbered]{algorithm2e}
\usepackage{tablefootnote}

\usepackage{hyperref}
\usepackage{xurl}

\DeclareMathOperator*{\sE}{\mathbb{E}}

\def\cL{{\mathcal{L}}}

\usepackage[accepted]{icml2022}

\usepackage{amsmath}
\usepackage{amssymb}
\usepackage{mathtools}
\usepackage{amsthm}
\usepackage{pifont}

\usepackage{xurl}

\usepackage[capitalize,noabbrev]{cleveref}

\theoremstyle{plain}

\theoremstyle{definition}

\theoremstyle{remark}

\usepackage[textsize=tiny]{todonotes}

\icmltitlerunning{
 Removing Batch Normalization Boosts Adversarial Training
}

\begin{document}

\twocolumn[
\icmltitle{
Removing Batch Normalization Boosts Adversarial Training
}



\icmlsetsymbol{equal}{*}

\begin{icmlauthorlist}
\icmlauthor{Haotao Wang}{utaustin}
\icmlauthor{Aston Zhang}{amazon}
\icmlauthor{Shuai Zheng}{amazon}
\icmlauthor{Xingjian Shi}{amazon}
\icmlauthor{Mu Li}{amazon}
\icmlauthor{Zhangyang Wang}{utaustin}
\end{icmlauthorlist}

\icmlaffiliation{utaustin}{University of Texas at Austin, Austin, USA}
\icmlaffiliation{amazon}{Amazon Web Services, Santa Clara, USA}

\icmlcorrespondingauthor{Haotao Wang}{\href{mailto:htwang@utexas.edu}{htwang@utexas.edu}}
\icmlcorrespondingauthor{Aston Zhang}{\href{mailto:astonz@amazon.com}{astonz@amazon.com}}
\icmlcorrespondingauthor{Zhangyang Wang}{\href{mailto:atlaswang@utexas.edu}{atlaswang@utexas.edu}}

\icmlkeywords{Machine Learning, ICML}

\vskip 0.3in
]



\printAffiliationsAndNotice{Work done during the first author's internship at Amazon Web Services.}  

\begin{abstract}

Adversarial training (AT) defends deep neural networks against adversarial attacks. One challenge that limits its practical application is the performance degradation on clean samples. 
A major bottleneck identified by previous works is the widely used batch normalization (BN), which struggles to model the different statistics of clean and adversarial training samples in AT. 
Although the dominant approach is to extend BN to capture this mixture of distribution, we propose to completely eliminate this bottleneck by removing all BN layers in AT. 
Our \textbf{no}rmalizer-\textbf{f}ree \textbf{ro}bu\textbf{s}t \textbf{t}raining (NoFrost) method extends recent advances in normalizer-free networks to AT for its unexplored advantage on handling the mixture distribution challenge. 
We show that NoFrost achieves adversarial robustness with only a minor sacrifice on clean sample accuracy. 
On ImageNet with ResNet50, NoFrost achieves $74.06\%$ clean accuracy, which drops merely $2.00\%$ from standard training.
In contrast, BN-based AT obtains $59.28\%$ clean accuracy, suffering a significant $16.78\%$ drop from standard training. 
In addition, NoFrost achieves a $23.56\%$ adversarial robustness against PGD attack, which improves the $13.57\%$ robustness in BN-based AT. 
We observe better model smoothness and larger decision margins from NoFrost, which make the models less sensitive to input perturbations and thus more robust. 
Moreover, when incorporating more data augmentations into NoFrost, it achieves comprehensive robustness against multiple distribution shifts.  
Code and pre-trained models are public\footnote{\url{https://github.com/amazon-research/normalizer-free-robust-training}}. 
\end{abstract}

\section{Introduction} \label{sec:intro}
Deep neural networks (DNNs) are vulnerable to adversarial attacks \cite{szegedy2013intriguing}, which generate adversarial images by adding slight manipulations on the original images to falsify model predictions. 
One of the most effective methods to defend adversarial attacks is adversarial training (AT)~\cite{goodfellow2014explaining,zhang2019theoretically}. 
It jointly fits a model on clean (original) images and adversarial images to improve the model's adversarial robustness (i.e., the accuracy on adversarial images). 
An improved adversarial robustness often comes at the cost of reducing accuracy on clean samples~\cite{madry2017towards,tsipras2018robustness,ilyas2019adversarial}. 
However, for many real-world applications, \textit{high clean accuracy} is a basic requirement while adversarial robustness is a favorable bonus.
It is hence desirable to sustain high clean accuracy while achieving high adversarial robustness.

Previous works \cite{xie2019intriguing,xie2019adversarial} pointed out that the widely used batch normalization (BN) layer \cite{ioffe2015batch} contributes to the undesirable trade-off between clean accuracy and adversarial robustness.
In AT, clean and adversarial images are drawn from two different distributions. It is  challenging for BN to capture those two different normalization statistics. \citet{xie2019intriguing} proposed mixture BN (MBN) strategy for AT, which routes clean and adversarial images through two separate BN paths. 
MBN is adopted as the default option for many follow-up works \cite{xie2019intriguing,xie2019adversarial,merchant2020does,li2020shape,wang2020once,wang2021augmax}. 
But as pointed out by the authors, it faces practical limitations: In practice there is no oracle to tell which BN path to choose during inference for each test sample.

We explore an alternative solution. Since BN has limited capacity to estimate normalization statistics of samples from heterogeneous distributions, can we  
better handle this mixture distribution challenge by removing all BN layers in AT? 
Replacing BN with other normalization layers that do not calculate statistics across samples, such as instance normalization (IN) \cite{ulyanov2016instance}, brings a small amount of benefit, but the results are still unsatisfying as shown in our experiments (\autoref{tab:in}). 
Instead, 
we focus on recently proposed normalizer-free networks~\cite{brock2021characterizing, brock2021high}. 
Although these networks were proposed to match state-of-the-art accuracy on standard training with improved hardware performance and memory efficiency, we leverage them for their unexplored benefit to handle data from mixture distributions. 

To this end, we propose the \textbf{no}rmalizer-\textbf{f}ree \textbf{ro}bu\textbf{s}t \textbf{t}raining (NoFrost) method to improve the trade-off between clean accuracy and adversarial robustness for AT. 
NoFrost is based on NF-ResNet~\cite{brock2021characterizing}, a ResNet~\cite{he2016deep} variant without normalization layers but achieving an comparable accuracy on ImageNet~\cite{deng2009imagenet}. Experimental results show that NoFrost achieves a better accuracy-robustness trade-off compared with previous state-of-the-art AT methods based on BN or MBN models.  
For example, on ImageNet with the ResNet50 backbone, NoFrost achieves $11.96\%$ higher robustness against APGD-CE attack \cite{croce2020reliable} and $0.42\%$  higher clean accuracy {simultaneously} compared with MBNAT \cite{xie2019intriguing}.
NoFrost also achieves $12.15\%$ higher clean accuracy and $7.25\%$ higher robustness against APGD-CE attack {simultaneously} than TRADES-FAT \cite{zhang2020attacks}.
To explain the effectiveness of NoFrost, we demonstrate that NoFrost has better model smoothness \cite{zhang2019theoretically} and larger decision margins \cite{yang2020boundary}.

Moreover, we show that NoFrost can be generalized towards \emph{comprehensive robustness} against distribution shifts beyond adversarial samples. 
In particular, we jointly train models on images generated by adversarial attacks and two other robust data augmentation methods, namely DeepAugment \cite{hendrycks2020many} and texture-debiased augmentation (TDA) \cite{hermann2020origins}. This extended version of NoFrost simultaneously achieves better or comparable adversarial robustness and accuracy on multiple out-of-distribution (OOD) benchmark datasets (i.e., OOD robustness), such as ImageNet-C \cite{hendrycks2019benchmarking}, Imagenet-R \cite{hendrycks2020many}, and ImageNet-Sketch \cite{wang2019learning}, compared with previous state-of-the-art robust learning methods including DeepAugment and texture-debiased augmentation.

In summary, our contributions are as follows:
\begin{enumerate}
    \item We propose NoFrost to improve the clean accuracy and adversarial robustness trade-off in AT. Our approach is simple and straightforward: just removing all BN layers to address the mixture distribution challenge.
    \item To the best of our knowledge, we for the first time apply normalizer-free models to AT. We demonstrate the unexplored advantage of normalizer-free models in handling data from mixture distributions.
    \item 
    We show that NoFrost achieves substantially better accuracy-robustness trade-off on ImageNet. 
    Using NF-ResNet50, NoFrost achieves $74.06\%$ clean accuracy, dropping merely $2.00\%$  from standard training, plus $23.56\%$ adversarial robustness against PGD attack.
    In comparison, BN-based AT obtains only $59.28\%$ clean accuracy with $13.57\%$ adversarial robustness using ResNet50. 
    \item We demonstrate that when combining adversarial samples with other data augmentation methods, NoFrost can simultaneously achieve adversarial robustness and out-of-distribution robustness.
\end{enumerate}

\section{Preliminary}
\label{sec:preliminary}

Before diving deep into the hidden benefit on model robustness brought by normalizer-free training,
we need to describe the concept of model robustness that includes both adversarial robustness and OOD robustness (Section \ref{sec:at}).
Then we will revisit the mixture distribution challenge (Section \ref{sec:mixture-distribution-challenge}) and normalizer-free networks (Section \ref{sec:related-works}).

\subsection{Model Robustness} \label{sec:at}

Model robustness refers to a model's performance under various data distribution shifts. Here we review two distribution shifts, adversarial examples and out-of-distribution examples, which are related to our work, and methods for improving model robustness. 

Adversarial robustness refers to a model's performance on adversarial samples. These adversarial samples are  modified from the original (clean) samples by adversarial attacks \cite{szegedy2013intriguing,madry2017towards,carlini2017towards,xiao2018spatially,croce2020reliable} to falsify the model. 
To defend these attacks, one of the most effective defense methods is adversarial training (AT)~\cite{goodfellow2014explaining,madry2017towards,zhang2019theoretically,zhang2020attacks, xie2019intriguing}. 
AT trains a model on both clean and adversarial images.  
Given a pair of clean image $\vx$ and its label $y$ sampled from the data distribution $\mathcal D$, 
AT learns a robust classifier $f_\vtheta$ with parameters $\vtheta$ by
\begin{equation}
\begin{split}
    \min_{\vtheta} \sE_{(\vx,y) \sim \mathcal{D}} ~ (1-\lambda)\cL(f_\vtheta(\vx),y) + 
    \lambda\cL(f_\vtheta(\vx^*),y), 
\label{eq:loss}
\end{split}
\end{equation}
where $\cL(\cdot,\cdot)$ is the cross-entropy loss function. 
The adversarial image $\vx^*$ in \autoref{eq:loss} is generated from $\vx$ by PGD attack \cite{madry2017towards}:
\begin{equation*}
\begin{split}
&\vx^{(0)} = \text{RandomSample}(\mathcal{B}(\vx,\eps)),\\
&\vx^{(t+1)} = \Pi_{\mathcal{B}(\vx,\eps)}(\vx^{(t)} + \alpha \cdot \text{sign}(\nabla_{\vx^{(t)}} \cL(f_\vtheta(\vx^{(t)}),y))), \\
&\vx^* = \vx^{(T)}, \\
\end{split}
\end{equation*}
where $\mathcal{B}(\vx,\eps)$ is the $\ell_{\infty}$ ball with radius $\eps$ around $\vx$, the initialization $\vx^{(0)}$ is randomly sampled from $\mathcal{B}(\vx,\eps)$, $\Pi_{\mathcal{B}(\vx,\eps)}$ means the nearest projection to $\mathcal{B}(\vx,\eps)$, 
$T$ is the total number of iterations, and $\alpha$ is the step size.
The hyper-parameter $\lambda$ controls the weight of the loss on adversarial samples. 
When $\lambda=0$ we obtain standard training.   
When $\lambda=1$ we obtain PGDAT \cite{madry2017towards}.
Some previous works set $\lambda=0.5$ \cite{kurakin2018adversarial,wang2020once} to  trade off clean accuracy and adversarial robustness, which we denote as standard adversarial training (SAT). 
Other works improve the above training approach. 
For example, TRADES \cite{zhang2019theoretically} simultaneously optimizes classification error and model smoothness. 
FAT \cite{zhang2020attacks} uses early-stopped PGD attack to generate ``friendly'' adversarial samples, which can also be combined with TRADES (termed as TRADES-FAT). 

Out-of-distribution robustness refers to the model's performance on out-of-distribution (OOD) examples. 
To evaluate OOD robustness, there exists multiple benchmark datasets. 
ImageNet-C~\cite{hendrycks2019benchmarking} adds natural corruptions such as Gaussian noise and  motion blur on the ImageNet validation set.
ImageNet-Sketch~\cite{wang2019learning} contains  sketch-like images to evaluate cross-domain transferability. 
ImageNet-A~\cite{hendrycks2019natural} contains naturally occurring real-world images which falsify state-of-the-art image classifiers.
ImageNet-R~\cite{hendrycks2020many} contains renditions such as painting and sculpture.  
To improve OOD robustness, a popular approach is through data augmentations~\cite{zhong2017random,cubuk2018autoaugment,geirhos2018,yun2019cutmix,wang2019learning,hendrycks2020many,gong2020maxup,wang2021augmax}.
For example, DeepAugment~\cite{hendrycks2020many} first adds random noise onto the weights of an image-to-image model (e.g., an image super-resolution model).
It then feeds clean images to the noisy image-to-image model and uses the output images as augmented data.
In this way, DeepAugment obtains diverse augmented images and thus achieves state-of-the-art robustness on ImageNet-C and ImageNet-R. 
Texture-debiased augmentation (TDA) \cite{hermann2020origins} stacks color distortion, less aggressive random crops, and other simple augmentations to debias a model towards textures and is shown to improve model generalizability (e.g., from ImageNet to ImageNet-Sketch).

\begin{figure}[t] 
	\centering
	\includegraphics[width=0.9\linewidth]{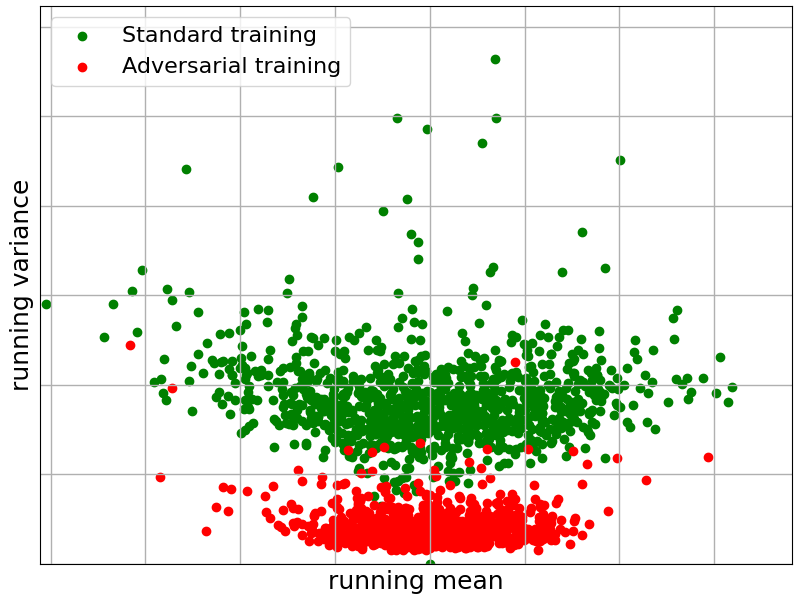} 
	\vspace{-1em}
	\caption{The mixture distribution challenge in adversarial training.
	Specifically, we show the channel-wise BN statistics of the $15$-th layer in ResNet26 models obtained by standard training and adversarial training, respectively. 
	Each dot represents the running mean and variance of a channel in the BN layer.
	We can see that clean and adversarial training samples have different feature statistics, and thus are sampled from different underlying distributions. Similar observations were first made in \cite{xie2019intriguing}.
	}
	\label{fig:intuition}
\end{figure}

\subsection{The Mixture Distribution Challenge}
\label{sec:mixture-distribution-challenge}

\citet{xie2019intriguing} have shown that solving \autoref{eq:loss} with traditional BN-based networks leads to unsatisfying trade-off between clean accuracy and adversarial robustness. 
The underlying reason is that clean and adversarial images are sampled from different distributions. It is difficult for BN to estimate the correct normalization statistics of such mixture of distributions.
We show the misalignment between clean and adversarial distributions in \autoref{fig:intuition}.\footnote{Similar observations were first made in \cite{xie2019intriguing}. We show them here for a more self-contained introduction.}
Specifically, we train two ResNet26 models on ImageNet by only using clean images (i.e., $\lambda=0$ in \autoref{eq:loss}) and adversarial images (i.e., $\lambda=1$), respectively.
We then plot the channel-wise BN statistics of the $15$-th layer (other layers are similar) in both models. 
As we can see, the running means and variances  of clean (the green dots) and adversarial images (the red dots) are significantly different.

To solve this mixture distribution challenge, \citet{xie2019intriguing} proposed MBNAT. It  uses a mixture BN (MBN) strategy to disentangle clean and adversarial statistics. 
Specifically, it uses two parallel BNs in each normalization layer, 
denoted by $\text{BN}_c$ and $\text{BN}_a$. 
During training, clean (adversarial) images are routed to $\text{BN}_c$ ($\text{BN}_a$).
As a result, $\text{BN}_c$ ($\text{BN}_a$) only estimates the distribution of clean (adversarial) images to avoid modeling a mixture distribution.

During inference, we should use $\text{BN}_c$ ($\text{BN}_a$) for a clean (adversarial) test image. There is, however, no oracle to tell us whether a test image is clean or adversarial. As shown by \citet{xie2019intriguing,xie2019adversarial}, it is difficult to choose which path in practise: If $\text{BN}_c$ is used, the model will have a good clean accuracy while sacrificing robustness, and vice versa 
(see Appendix \ref{sec:appx-mixbn} for details).

\subsection{Normalizer-Free Networks} \label{sec:related-works}
Batch normalization (BN) is originally proposed as a regularization to enable stable training of DNNs \cite{ioffe2015batch}, and is then adopted as a basic building block of DNNs.
Research on normalizer-free networks aims to remove BN from DNNs for better hardware efficiency.
The first attempt to train normalizer-free (NF) deep residual networks 
uses stable weight initialization methods \cite{zhang2018fixup,de2020batch,bachlechner2020rezero}.
For example, SkipInit initializes residual blocks in NF networks close to identity mappings, ensuring signal propagation and well-behaved gradients \cite{de2020batch}.
Although these initialization methods enable stable training of NF deep residual networks, the obtained test accuracy is still lower than that of well-tuned normalized models. 
More recently, \citet{brock2021characterizing} first obtained NF networks with performance competitive with traditional BN-based ResNets \cite{he2016deep} and EfficientNets \cite{tan2019efficientnet}. 
The authors proposed scaled weight standardization which normalizes the weights in each layer to prevent mean shift in hidden activations and thus stabled the training.
Given the weight matrix $\mW$ of a convolutional or fully connected layer, the proposed scaled weight standardization takes the following form:
\begin{equation*}
    \hat{\mW}_{i,j}=\gamma\frac{\mW_{i,j}-\mu_i}{\sigma_i\sqrt{N}},
\end{equation*}
where $\mu_i$ and $\sigma_i$ are the mean and standard deviation of the $i$-th row of $\mW$,  $\gamma$ is a fixed number, and $N$ is the batch size.
This constraint is imposed throughout training as a differentiable operation in the forward pass.
\citet{brock2021high} further proposed an adaptive gradient clipping method, which enables normalizer-free models to train with large batch sizes and strong data augmentations for better test accuracy.

\section{Method}
\label{sec:nfat}

\subsection{NoFrost for Adversarial Training}

We adopt a simple strategy to address the mixture distribution challenge in AT.
Since this challenge arises from the limited capability of BN to simultaneously encode the heterogeneous distributions of clean and adversarial samples, we simply remove all BN layers in the model.
Specifically, our normalizer-free robust training (NoFrost) method, solves the AT problem in \autoref{eq:loss} by using NF networks \cite{brock2021characterizing} as the model $f_{\vtheta}$. 
We use PGD attack \cite{madry2017towards} to generate adversarial samples in NoFrost,
and set $\lambda=0.5$ for simplicity\footnote{Searching for the optimal $\lambda$ value may lead to better performance, which we leave for future work.}.
Without BN layers, NoFrost naturally overcomes the limitation of inference-time oracle BN selection in MBNAT described in Section~\ref{sec:mixture-distribution-challenge}.

\subsection{NoFrost$^*$ for Comprehensive Robustness}
\label{subsec:nofrost-star}

To achieve the more challenging \emph{comprehensive robustness} (i.e., to be simultaneously robust against multiple adversarial attacks and naturally occurring distribution shifts), we further generalize NoFrost by combining it with other robust data augmentation methods. 
Different robust data augmentations have different strength and weakness: the best method against one type of distribution shift may not be the best against another. 
For example, AT uses adversarial attack as a data augmentation method that augments clean images by adding worst-case additive noises, and achieves state-of-the-art robustness against adversarial attacks; 
however, AT cannot achieve state-of-the-art robustness against natural distribution shifts (e.g., in ImageNet-R and ImageNet-Sketch).
Similarly, DeepAugment is the state-of-the-art data augmentation method against distribution shifts caused by different renditions (e.g., art, cartoons, and graffiti in ImageNet-R), but has little benefit against adversarial attacks.

As a result, a naive way to achieve comprehensive robustness against multiple types of distribution shifts is to combine multiple robust data augmentations during training.
However, jointly training on multiple different data augmentations may cause an even harder mixture distribution challenge than adversarial training that uses only one augmentation (i.e., adversarial attack).

In view of this, we extend NoFrost by incorporating more data augmentation methods to achieve comprehensive robustness.
By default, we add two new robust data augmentation methods, namely DeepAugment and texture-debiased augmentation (TDA), for the extended version of NoFrost (denoted as NoFrost$^*$). 
Formally, the optimization problem of NoFrost$^*$ is:
\begin{equation}
\begin{split}
    \min_{\vtheta} \sE_{(\vx,y) \sim \mathcal{D}} ~ (&\cL(f_\vtheta(\vx),y) + \\
    &\cL(f_\vtheta(\hat{\vx}),y) + \cL(f_\vtheta(\vx^*),y))/3, 
\end{split}
\end{equation}
where $\hat{\vx}$ is the augmented image generated from $\vx$ using either DeepAugment or TDA (each with half probability), $\vx^*$ is the adversarial image generated from $\vx$, and $f_\vtheta$ is a normalizer-free network. Other notations have the same meaning as in \autoref{eq:loss}.

\begin{figure*}[ht]
\centering
\setlength{\tabcolsep}{1pt}
	\begin{tabular}{cccc}
    	\includegraphics[width=0.245\linewidth]{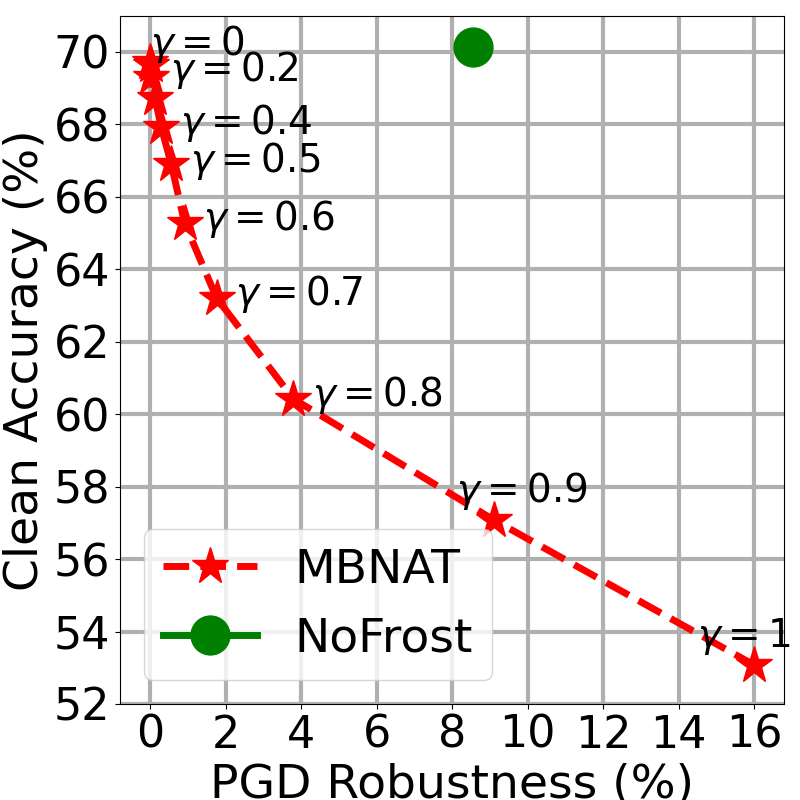} & 
    	\includegraphics[width=0.245\linewidth]{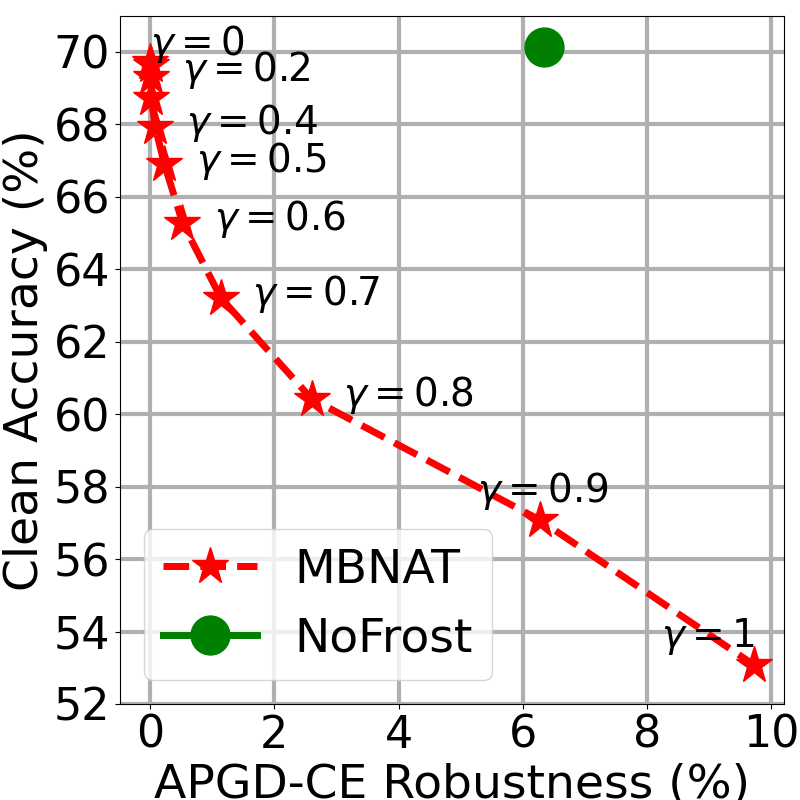} &
    	\unskip\hfill{\vrule width 1pt}\hfill
    	\includegraphics[width=0.245\linewidth]{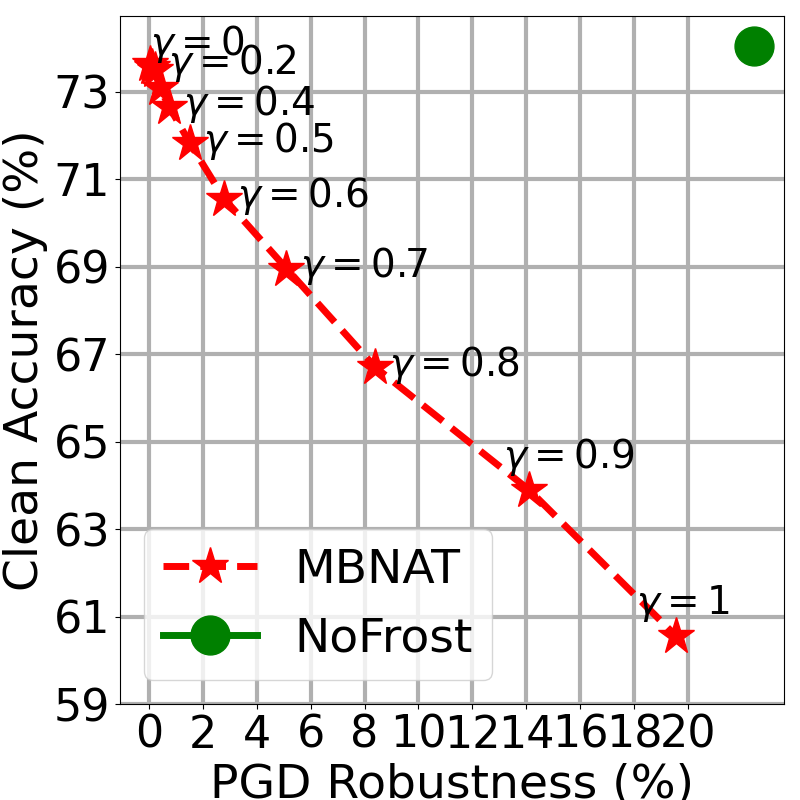} & 
    	\includegraphics[width=0.245\linewidth]{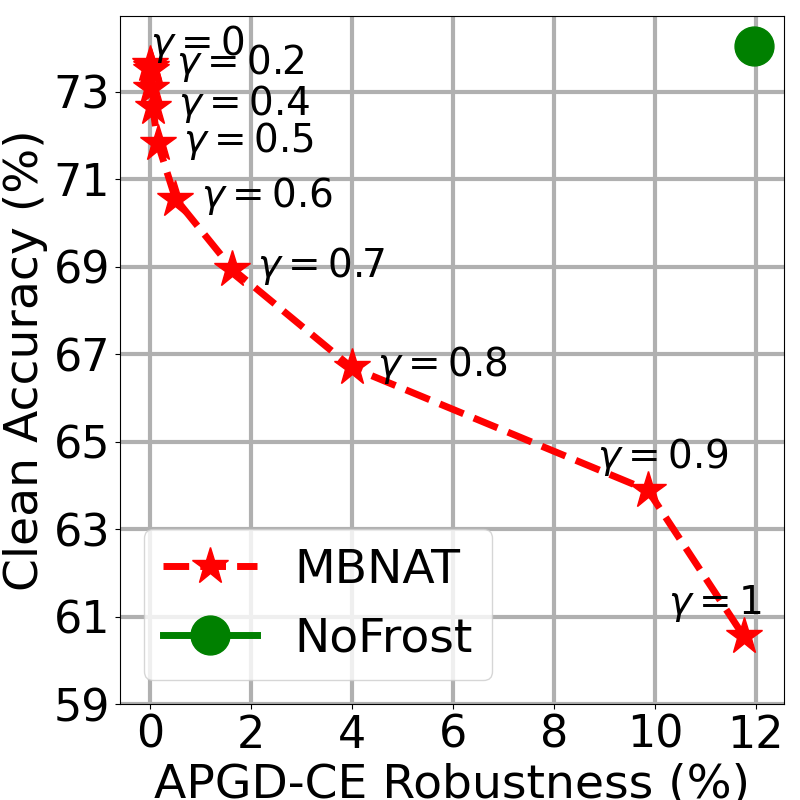} \\
    	\includegraphics[width=0.245\linewidth]{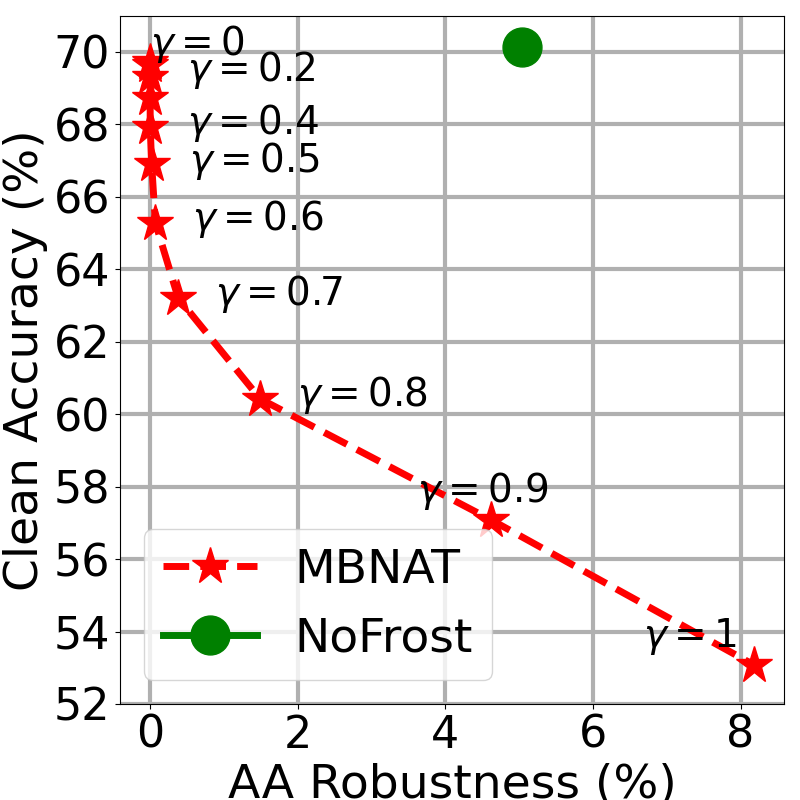} & 
    	\includegraphics[width=0.245\linewidth]{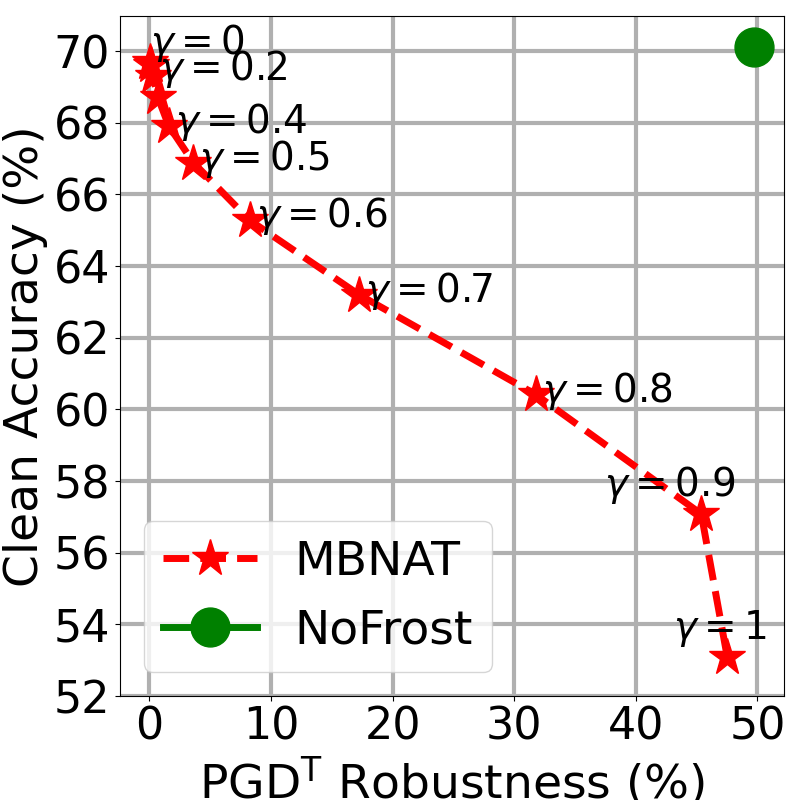} & 
    	\unskip\hfill{\vrule width 1pt}\hfill
    	\includegraphics[width=0.245\linewidth]{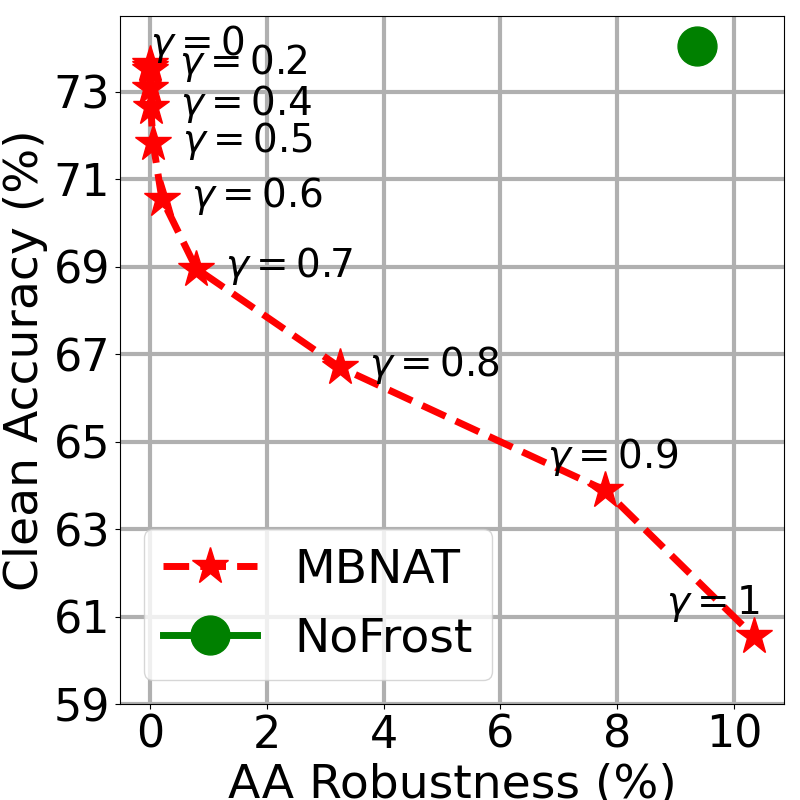} &
    	\includegraphics[width=0.245\linewidth]{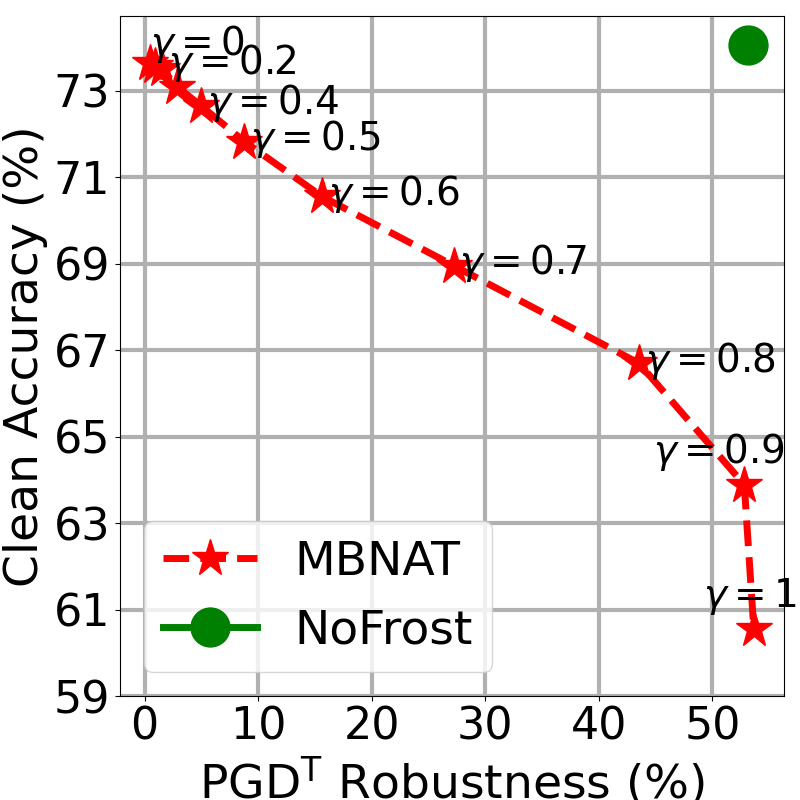} \\
    	\multicolumn{2}{c}{(a) ResNet26} & \multicolumn{2}{c}{(b) ResNet50}
	\end{tabular}
\caption{Trade-off between robustness and accuracy on ResNet26 (sub-figure (a)) and ResNet50 (sub-figure (b)) trained by MBNAT and NoFrost. Adversarial robustness is evaluated on PGD (the top-left panel in each sub-figure), APGD-CE (the top-right panel in each sub-figure),
AutoAttack (AA, the bottom-left panel in each sub-figure) and targeted PGD (denoted as $\text{PGD}^{\text{T}}$; the bottom-right panel in each sub-figure) attacks. 
$\gamma$ is the weight for interpolation between the two BNs in MBNAT.
}
\label{fig:tradeoff_26}
\end{figure*}

\section{Experiments}
\label{sec:exp}
In this section, we first describe the general experimental settings (Section \ref{sec:settings}).
Then we show that NoFrost outperforms previous adversarial training methods (Section \ref{sec:nfat-results}) and achieves nicer properties such as better model smoothness and larger decision margins (Section \ref{sec:visualization}). 
Finally, we evaluate NoFrost$^*$ for comprehensive robustness against multiple distribution shifts (Section \ref{sec:crt-results}).

\subsection{Experimental Settings}
\label{sec:settings}

\paragraph{Datasets, models, and metrics}
All methods are trained on the ImageNet \cite{deng2009imagenet} dataset. 
We use ResNet26 and ResNet50 \cite{he2016deep} backbones, with different normalization strategies: BN \cite{ioffe2015batch}, MBN \cite{xie2019intriguing}, and NF \cite{brock2021characterizing}.
We evaluate \textit{clean accuracy} on the ImageNet validation set, and use the accuracy on adversarial test images as a metric for \emph{adversarial robustness}.
We generate adversarial images on the ImageNet validation set using white-box attacks (PGD~\cite{madry2017towards}, APGD-CE \cite{croce2020reliable}, APGD-DLR \cite{croce2020reliable}, MIA \cite{dong2018boosting}, CW \cite{carlini2017towards}\footnote{We use the $\ell_\infty$ version of CW attack following \cite{zhang2020attacks}.}),  black-box attacks (RayS \cite{chen2020rays}, and Square \cite{andriushchenko2020square}) and the AutoAttack (AA) \cite{croce2020reliable}.
We evaluate \textit{OOD robustness} against naturally occurring distribution shifts by measuring accuracy on ImageNet-C \cite{hendrycks2019benchmarking}, ImageNet-R \cite{hendrycks2020many}, and ImageNet-Sketch \cite{wang2019learning}.

\vspace{-0.5em}
\paragraph{General hyper-parameters}
\label{sec:appx-hyper}
For all experiments, we train on ImageNet for $90$ epochs. We use the SGD optimizer with momentum $0.9$. Batch size is $256$. Weight decay factor is $5\times10^{-5}$. The initial learning rate is 0.1 and decays following a cosine annealing scheduler.
All experiments are conducted with 8 NVIDIA V100 GPUs. 

\vspace{-0.5em}
\paragraph{Implementation of adversarial training}
We study adversarial robustness under perturbation magnitude $\eps=8$ (on scale of 0 to 255 in unsigned 8-bit pixels).
We set the maximum PGD attack iteration number $T=10$ during training for all adversarial training methods. 
For TRADES and TRADES-FAT, we set the loss trade-off hyper-parameter to $1$ following the original TRADES paper \cite{zhang2019theoretically}.
For FAT and TRADES-FAT, we set the PGD early-stop iteration to $1$ following the original FAT paper \cite{zhang2020attacks}.

\vspace{-0.5em}
\paragraph{Details for robustness evaluation}
Following \cite{zhang2019theoretically,zhang2020attacks}, we set attack iterations to be $20$ for all white-box attacks. 
For all black-box attacks, we allow $400$ queries per-sample on all compared models. 
For AutoAttack, we use the fast version with APGD-CE and APGD-DLR attacks.
For RayS, we evaluate on a subset of ImageNet validation set with $1000$ images due to high computational costs.

\subsection{Adversarial Training Results}
\label{sec:nfat-results}

\begin{table*}[ht]
\centering
\caption{Adversarial robustness of ResNet26 under perturbation magnitude $\eps=8$. Classification accuracy on clean images and under different adversarial attacks are reported. The best and second to the best numbers are shown in bold and underlined, respectively.}
\setlength{\tabcolsep}{10pt}
\resizebox{0.95\linewidth}{!}{
\begin{tabular}{c|c|ccccc|cc|c}
\toprule
\multirow{2}{*}{\textbf{Method}} & \multirow{2}{*}{\textbf{Clean}} & \multicolumn{5}{c|}{\textbf{White-box Attacks}} & \multicolumn{2}{c|}{\textbf{Black-box Attacks}} & \multirow{2}{*}{\textbf{AA}} \\
&  & \textbf{PGD} & \textbf{APGD-CE} & \textbf{APGD-DLR} & \textbf{MIA} & \textbf{CW} & \textbf{RayS} & \textbf{Square} & \\
\midrule
ST & \textbf{72.68} &  0.01 & 0.00 & 0.00 & 0.00 & 0.00 & 18.2 & 27.5 & 0.00 \\
\midrule
SAT & 52.65 & 10.55 & 5.02 & 5.30 & 8.84 & 9.18 & 30.5 & 44.7 & 3.78 \\
\midrule
TRADES & 39.64 & 9.94 & \underline{6.24} & 4.02 & 8.33 & 6.37 & 20.7 & 32.8 & 3.54 \\
\midrule
FAT  & 58.72 & 6.97 & 2.35 & 2.68 & 6.59 & 6.37 & \underline{33.6} &  \textbf{50.8} & 1.70 \\
\midrule
TRADES-FAT & 55.65 & \underline{11.91} & {5.79} & \underline{6.14} & \underline{10.83} & \textbf{10.81} & 31.1 & 46.7 & \underline{4.63} \\
\midrule 
NoFrost & \underline{70.13} & \textbf{12.24} & \textbf{6.34} & \textbf{6.60} & \textbf{21.83} & \underline{10.18} & \textbf{34.5} & \underline{48.3} & \textbf{5.04} \\
\bottomrule
\end{tabular}
}
\label{tab:resnet26}
\end{table*}

\begin{table*}[htb]
\centering
\caption{Adversarial robustness of ResNet50 under perturbation magnitude $\eps=8$. Classification accuracy on clean images and under different adversarial attacks are reported. The best and second to the best numbers are shown in bold and underlined, respectively.}
\setlength{\tabcolsep}{10pt}
\resizebox{0.95\linewidth}{!}{
\begin{tabular}{c|c|ccccc|cc|c}
\toprule
\multirow{2}{*}{\textbf{Method}} & \multirow{2}{*}{\textbf{Clean}} & \multicolumn{5}{c|}{\textbf{White-box attacks}} & \multicolumn{2}{c|}{\textbf{Black-box attacks}} & \multirow{2}{*}{\textbf{AA}}\\
&  & \textbf{PGD} & \textbf{APGD-CE}  & \textbf{APGD-DLR}  & \textbf{MIA} & \textbf{CW} & \textbf{RayS} & \textbf{Square}\\
\midrule
ST & \textbf{76.06} & 0.04 & 0.00 & 0.00 & 0.00 & 0.00 & 22.5 & 31.4 & 0.00 \\
\midrule
SAT & 59.28 & 13.57 & 7.80 & \underline{8.46} & 10.28 & 11.02 & 27.4 & 40.2 & 6.23 \\
\midrule 
TRADES & 49.25 & \underline{14.80} & \underline{9.20} & 8.19 & \underline{12.97} & 11.80 & 32.6 & 39.5 & \underline{6.66} \\
\midrule 
FAT & 58.94 &  12.45 & 5.48 & 7.16 & 12.56 & \underline{12.24} & \underline{35.9} &    \textbf{51.4} & 4.73 \\
\midrule 
TRADES-FAT & 60.52 & 11.67 & 4.71 & 5.90 & 11.28 & 10.29 & 34.5 & \underline{48.6} & 3.87 \\
\midrule 
NoFrost & \underline{74.06} & \textbf{22.45} & \textbf{11.96} & \textbf{13.37} & \textbf{36.11} &  \textbf{19.17} & \textbf{36.1} & {43.1} & \textbf{9.36} \\
\bottomrule
\end{tabular}
}
\label{tab:resnet50}
\end{table*}

We first compare NoFrost with MBNAT, the \textit{de facto} solution for resolving the AT mixture distribution challenge \cite{xie2019intriguing,xie2019adversarial,merchant2020does,li2020shape,wang2020once,wang2021augmax,wang2022partial}.
As discussed in Section \ref{sec:mixture-distribution-challenge}, MBN requires an empirical weighting value $\gamma$ to be set for interpolation between $\text{BN}_c$ and $\text{BN}_a$ during inference (see Appendix~\ref{sec:appx-mixbn} for more details). 
The original MBNAT paper \cite{xie2019intriguing} uses $\gamma=1$ to pursue the best adversarial robustness. In another work by the same first author \cite{xie2019adversarial}, $\gamma=0$ is applied for the best clean accuracy. 
We uniformly sample $\gamma$ from interval $[0,1]$ to obtain the robustness-accuracy Parato frontier of MBNAT.

Comparison results between NoFrost and MBNAT on ResNet26 and ResNet50 are shown in \autoref{fig:tradeoff_26}.
A point closer to the top-right corner represents a more desired model with higher clean accuracy and adversarial robustness. 
For MBNAT models, as the value of $\gamma$ increases from $0$ to $1$, the influence of $\text{BN}_a$ gradually outweighs that of $\text{BN}_c$ (see Appendix \ref{sec:appx-mixbn} for more details). 
As a result, the adversarial robustness increases while the clean accuracy sharply drops. 
In contrast, NoFrost simultaneously achieves decent clean accuracy and adversarial robustness.  
In other words, NoFrost achieves a much more desired trade-off between clean accuracy and adversarial robustness compared with MBNAT.
For example, on ResNet26, NoFrost achieves 70.13\% accuracy and 6.34\% robustness against the APGD-CE attack. To achieve comparable accuracy, MBNAT needs to set $\gamma=0$ which leads to 0 robustness against APGD-CE attack (6.34\% less than NoFrost) and 69.71\% accuracy (0.42\% less than NoFrost).
On the other hand, to achieve comparable robustness with NoFrost, MBNAT needs to set $\gamma=0.9$ which leads to 57.08\% accuracy (13.05\% less than NoFrost) and 6.27\% robustness (0.07\% less than NoFrost).

We further compare NoFrost with other adversarial training methods, including TRADES, FAT, and TRADES-FAT, on ImageNet.
The results on ResNet26 and ResNet50 are shown in Table \ref{tab:resnet26} and \ref{tab:resnet50}, respectively.
NoFrost achieves significantly higher accuracy on clean images and better or comparable robustness against different attacks, compared with all those adversarial training methods.
For example, on ResNet26, NoFrost outperforms TRADES-FAT by 14.48\% on clean accuracy, and 0.55\% against APGD-CE attack.

\begin{table}[t]
\centering
\caption{Standard adversarial training (SAT) with IN-based networks yields worse robustness than NoFrost. Experiments conducted on ResNet26 with different normalizers.}
\resizebox{0.55\linewidth}{!}{
\begin{tabular}{ c|c|c }
\toprule
& \makecell{\textbf{Clean}} & \makecell{\textbf{PGD}}  \\
\midrule
SAT w/ BN & 52.65 & 10.55  \\
SAT w/ IN & \underline{56.78} & \underline{11.06} \\
NoFrost & \textbf{70.13} & \textbf{12.24}  \\
\bottomrule
\end{tabular}
}
\label{tab:in}
\end{table}

Since the mixture of distribution challenge is mainly caused by the limited capability of single BN layers to encode the mixture distribution of clean and adversarial samples, another possible solution is to replace BN with instance-level normalization layers, such as instance normalization (IN) \cite{ulyanov2016instance}. We denote the method of replacing BN with IN in SAT as ``SAT w/ IN". 
The results are shown in \autoref{tab:in}.
Both SAT w/ IN and NoFrost achieve better accuracy on clean images and robustness against PGD attack, compared with the naive BN counterpart (i.e., SAT).
This is intuitive since both methods are reasonable solutions for the mixture of distribution problem in adversarial training.
However, NoFrost achieves considerably better performance than SAT w/ IN, with 13.35\% higher accuracy and 1.18\% higher robustness against PGD attack.

\paragraph{Stability analyses}
In \autoref{tab:variance}, we show the stability analysis results on NoFrost over the  randomness in the algorithm (e.g., random initialization, random batch sampling). 
Specifically, we run NoFrost on ResNet26 with three different random seeds, and report the mean (denoted as $\mu$) and standard deviation (denoted as $\sigma$) of the testing results on those three models in the form of $\mu\pm\sigma$ in \autoref{tab:variance}.
We report accuracy on both clean and adversarial images generated by different attacks. 
As we can see, NoFrost has stable performance with small standard derivations on both clean accuracy and adversarial robustness.

\begin{table}[t]
\centering
\caption{Mean and standard deviation of NoFrost using ResNet26 with three different random seeds.}
\resizebox{1\linewidth}{!}{
\begin{tabular}{ c|c|c|c|c|c}
\toprule
& \makecell{\textbf{Clean}} & \makecell{\textbf{PGD}} & \makecell{\textbf{APGD-CE}} & \makecell{\textbf{APGD-DLR}} & \makecell{\textbf{AA}}  \\
\midrule
NoFrost & $70.15\pm0.03$ & $12.19\pm0.10$ & $6.34\pm0.04$ & $6.57\pm0.06$ & $5.01\pm0.09$ \\
\bottomrule
\end{tabular}
}
\label{tab:variance}
\end{table}

\subsection{NoFrost Leads to More Robust Model Properties}
\label{sec:visualization}

In this section, we show that NoFrost models have stronger model smoothness, larger decision margins \cite{wang2019improving,kim2021bridged}, and boundary thickness \cite{yang2020boundary}.
All these have been shown to benefit model robustness~\cite{ sanyal2020benign,wang2019improving,yang2020boundary}.
In the following, we provide definitions for these properties and empirically show how they are influenced by removing normalization layers in adversarial training.
\begin{itemize}[leftmargin=1em]
\item Decision margin: Following \cite{kim2021bridged}, we define $M(\vx)=\vp(\vx)_y-\max_{i \neq y}\vp(\vx)_i$ as the decision margin for a sample pair $(\vx,y)$, where $\vp(\vx)$ is the softmax probability of sample $\vx$. $M(\vx)<0$ indicates a wrong prediction on sample $\vx$.
\vspace{-0.5em}
\item Boundary thickness: Following \cite{yang2020boundary}, the boundary thickness of date $\vx$ is defined as
$T(\vx)=\|\vx-\vx^*\|_2\int_{0}^{1}\mathbb{I}\{\alpha<g_{ij}(t\vx+(1-t)\vx^*)<\beta\}dt$, where $g_{ij}(\cdot)=\vp(\cdot)_i-\vp(\cdot)_j$, $i$ and $j$ are the predicted labels of $\vx$ and $\vx^*$ respectively, and $\mathbb{I}\{\cdot\}$ is the indicator function.
It measures the distance between two level sets $g_{ij}(\cdot) = \alpha$ and $g_{ij}(\cdot) = \beta$ along the adversarial direction.
We set $\alpha=0$, $\beta=0.75$ and solve $\vx^*$ via a targeted 20-step PGD attack following the original paper \cite{yang2020boundary}.
\vspace{-0.5em}
\item Model smoothness: Following \cite{zhang2019theoretically,kim2021bridged}, we use the KL divergence $D(\vx)=\text{KL}(\vp(\vx)\|\vp(\vx^*))$ as a measurement for model smoothness for sample $\vx$, where $\vx^*$ is an adversarial image generated from $\vx$.
A smaller $D(\vx)$ indicates a stronger local model smoothness at $\vx$.
\end{itemize}
\vspace{-1em}

\begin{table}[ht]
\centering
\caption{Decision margin, boundary thickness, and model smoothness of adversarially trained (under $\eps=8$) ResNet26 models with different normalization strategies. The best and second-best values are bolded and underlined, respectively.}
\resizebox{0.9\linewidth}{!}{
\begin{tabular}{ c|c|c|c }
\toprule
\makecell{\textbf{Normalization}\\ \textbf{strategy} \\ \textbf{(Method)}} & \makecell{\textbf{Decision}\\ \textbf{margin} \\ \textbf{$M(\vx)$($\uparrow$)}} & \makecell{\textbf{Boundary}\\ \textbf{thickness} \\ \textbf{$T(\vx)$($\uparrow$)}} & \makecell{\textbf{Model}\\ \textbf{smoothness} \\ \textbf{$D(\vx)$($\downarrow$)}} \\
\midrule
BN (SAT) & \underline{0.3241} & \underline{17.51} & 4.927 \\  
MBN (MBNAT) & 0.3143 & 13.78 & \textbf{1.119} \\
NF (NoFrost) & \textbf{0.4700} & \textbf{31.49} & \underline{2.996} \\
\bottomrule
\end{tabular}
}
\label{tab:metrics}
\end{table}

We measure the above metrics on the 500 validation images from the first 10 classes on ImageNet. 
We report their mean values over the 500 images in \autoref{tab:metrics} and also show the distribution of those metrics using histograms in \autoref{fig:metrics} (in Appendix~\ref{sec:appx-hist}).
Compared with SAT, NoFrost leads to larger decision margins, thicker boundaries, and stronger model smoothness. 
All these three properties are beneficial for model robustness \cite{moosavi2019robustness,sanyal2020benign,wang2019improving,yang2020boundary}.

Another interesting observation is that, compared with SAT, MBNAT improves model smoothness while leaving the decision margin and boundary thickness almost unchanged.
In contrast, NoFrost improves all three properties over SAT. This is consistent with the recent finding that different defense methods improve robustness through different underlying mechanisms \cite{kim2021bridged}.
Our findings suggest that MBNAT improves model robustness mainly through improving model smoothness, while NoFrost improves robustness by simultaneously improving all three properties.

\subsection{Comprehensive Robustness}
\label{sec:crt-results}

\begin{figure}[htb]
\centering
\includegraphics[width=1\linewidth]{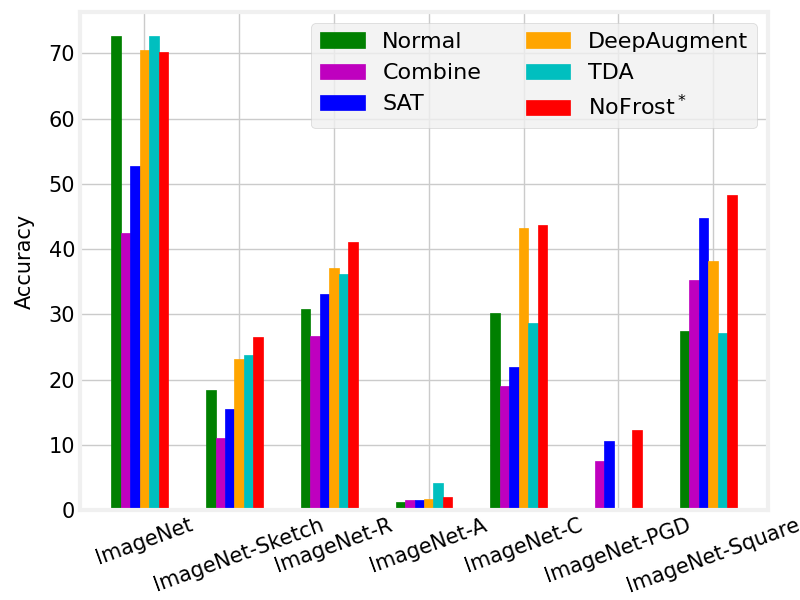}
\vspace{-1em}
\caption{Model performance (accuracy in percentage) on different benchmark datasets or adversarial attacks. All methods are trained on ImageNet with ResNet26.}
\label{fig:radar}
\end{figure}

\begin{figure}[htb]
\centering
\includegraphics[width=1\linewidth]{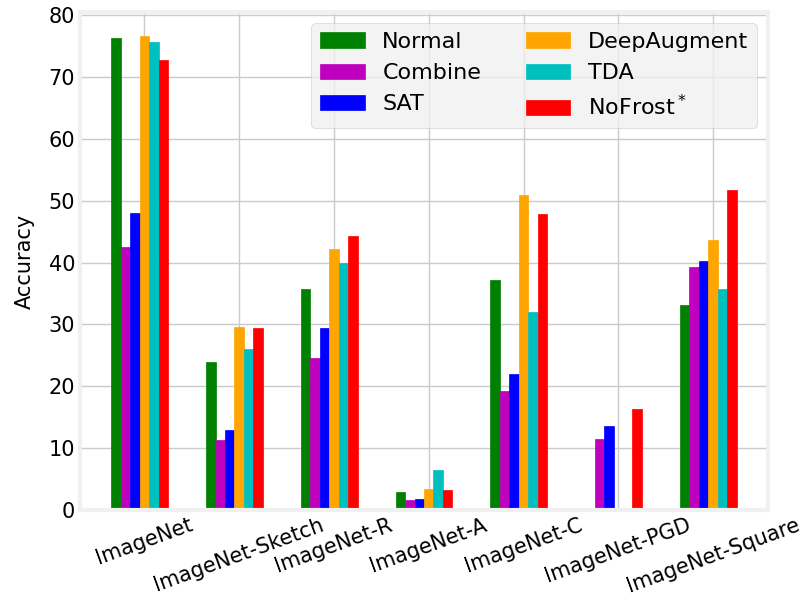}
\vspace{-1em}
\caption{Model performance (accuracy in percentage) on different benchmark datasets or adversarial attacks. All methods are trained on ImageNet with ResNet50.}
\vspace{-0.5em}
\label{fig:radar-rn50}
\end{figure}

Now we evaluate NoFrost$^*$ (\autoref{subsec:nofrost-star}) for comprehensive robustness  on three robustness benchmark datasets (ImageNet-C, ImageNet-R, and ImageNet-Sketch), two adversarial attacks (PGD and Square), together with clean accuracy on the ImageNet validation set.
Since NoFrost$^*$ jointly fits clean, adversarial, DeepAugment, and TDA samples, we compare it with the four stand-alone methods:
Standard training (training with only clean images), SAT (training with both clean and adversarial images), DeepAugment, and TDA.
We also include the naive combination of the four methods (i.e., jointly training on clean, adversarial, DeepAugment, and TDA samples on a traditional BN network) as a baseline, which is denoted as ``Combine''.
All methods are trained using the same settings in Section \ref{sec:settings}.
Results are shown in Figure \ref{fig:radar} and \ref{fig:radar-rn50}.

Notably, the naive combination performs the worst in most cases. This shows the inferent difficulty in fitting multiple heterogeneous augmentations under the traditional BN. 
In contrast, equipped with the new normalizer-free strategy, NoFrost$^*$ successfully fits all data augmentations within a single model and achieves comprehensive robustness. 
On ResNet26, NoFrost$^*$ achieves the best robustness on all evaluated OOD benchmark datasets and adversarial attacks. 
On ResNet50, although NoFrost$^*$ achieves slightly worse (3.06\% less) robustness on ImageNet-C than DeepAugment, it outperforms all baseline methods on other OOD benchmark datasets and adversarial attacks. 
For example, NoFrost$^*$ achieves 16.30\%, 8.20\%, and 2.05\% higher robustness than DeepAugment on PGD attack, Square attack, and ImageNet-R, respectively.

\section{Discussions}
Apart from this work and the MBN papers \cite{xie2019intriguing,xie2019adversarial}, there are other related works studying how BN affects model robustness. \citet{benz2021revisiting,schneider2020improving} proposed to improve model robustness against natural image corruptions (e.g., random Gaussian noise and motion blurring) by unsupervised model adaptation.
Specifically, they replace the BN statistic calculated on clean training images with those on unlabeled corrupted images.
AdvBN \cite{shu2021encoding} added adversarial perturbations on the BN statistics to increase model robustness against unseen distribution shifts such as style variations and image corruptions. 
\citet{galloway2019batch} and \citet{benz2021batch} observed that, in \emph{standard training}, BN grants models with better clean accuracy but harms their adversarial robustness. 
In contrast to their work, our paper utilizes normalizer-free networks to solve the mixture distribution challenge and improve the trade-off between clean accuracy and adversarial robustness in \emph{adversarial training}. 
More related works on machine learning robustness can be found in a recent survey paper \cite{mohseni2021practical}.

Our paper shows that removing BN can significantly boost adversarial training.
Yet, some existing test-time adaptation methods utilize the existence of BN to improve model robustness \cite{wang2020tent,nandy2021covariate,awais2020towards,benz2021revisiting}.   
Those methods are not directly applicable on normalizer-free networks, and thus NoFrost cannot be directly combined with those existing test-time adaptation methods for potentially further improved robustness. 
It will be our future work to study how to efficiently enable test-time adaption upon NoFrost, potentially by designing new test-time adaptation methods tailored for NF networks. 
On the other hand, NoFrost can potentially benefit from the future improvements in both fields of normalizer-free networks and adversarial training.   

\section{Conclusion}
In this paper, we address the issue of significant degradation on clean accuracy in adversarial training. The proposed NoFrost method removes all BNs in AT.  
NoFrost achieves a significantly more favorable trade-off between clean accuracy and adversarial robustness compared with previous BN-based AT methods: It achieves decent adversarial robustness with only minor degradation on clean accuracy. 
It is further generalized to achieve the more challenging goal of comprehensive robustness. 
We hope this study could be a stepping stone towards the exploration of normalizer-free training in improving model robustness and other fields with the challenge of data heterogeneity, such as distributed learning and domain generalization.

\vspace{-0.5em}
\section*{Acknowledgement}
\vspace{-0.5em}
Z.W. is supported by the U.S. Army Research Laboratory Cooperative Research Agreement W911NF17-2-0196 (IOBT REIGN) and an Amazon Research Award.

\bibliography{reference}
\bibliographystyle{icml2022}

\clearpage
\appendix


\section{How to Interpolate between Two BN Branches in MBNAT}
\label{sec:appx-mixbn}
We follow \cite{merchant2020does} to interpolate between the two BN branches (i.e., the $\text{BN}_c$ branch and $\text{BN}_a$ branch) \textit{during test time} for MBNAT.
Specifically, given an input test image $\vx$, we first forward it through the $\text{BN}_c$ branch (i.e., the MBN network using $\text{BN}_c$ at each normalization layer) to get the output logits $\vz_c$, and then through the $\text{BN}_a$ branch (i.e., the MBN network using $\text{BN}_a$ at each normalization layer) to get the output logits $\vz_a$.
We then average $\vz_c$ and $\vz_a$ with a weighting hyper-parameter $\gamma$, i.e.,  $\vz=(1-\gamma)\vz_c+\gamma\vz_a$.
Finally, we use the averaged logits $\vz$ as the input to the softmax function to get the final prediction probabilities. 
As a result, when the value of $\gamma$ increases from 0 to 1, the influence of $\text{BN}_a$ gradually outweighs that of $\text{BN}_c$. 
When $\gamma$ is $0$ or $1$, the MBN model falls back to the simple cases with only one BN ($\text{BN}_c$ when $\gamma=0$ or $\text{BN}_a$ when $\gamma=1$) at each normalization layer. 
This is the default interpolation method we used in our paper, which is denoted as ``MBNAT (logits)" or simply ``MBNAT" when used as the default.

Besides the one suggested in \cite{merchant2020does} (i.e., MBNAT (logits)), we have also investigated other possible interpolation methods between $\text{BN}_c$ and $\text{BN}_a$.
For example, we can interpolate the outputs of $\text{BN}_c$ and $\text{BN}_a$ at each MBN layer. 
Specifically, if the input feature of an MBN layer is denoted as $\vf_{i}$, then the output feature   $\vf_{o}=(1-\gamma)\text{BN}_c(\vf_{i})+\gamma\text{BN}_a(\vf_{i})$, where $\text{BN}_c(\cdot)$ and $\text{BN}_a(\cdot)$ are the batch normalization operations by $\text{BN}_c$ and $\text{BN}_a$, respectively.
We denote this method as ``MBNAT (all)".
We can also conduct this output mixing on some selected MBN layers, while keeping the two parallel outputs in other MBN layers. 
For example, we can randomly select $p\%$ MBN layers for mixing. We denote this method as ``MBNAT (random $p\%$)".

\begin{figure}[ht]
\centering
\includegraphics[width=0.7\linewidth]{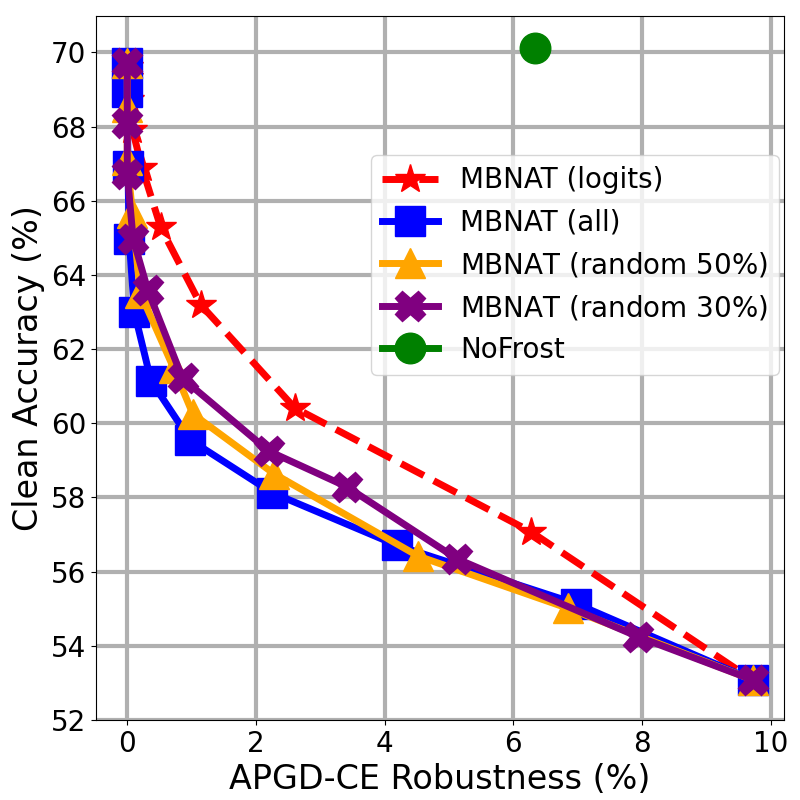}
\vspace{-1em}
\caption{Trade-off between robustness and accuracy of different interpolation strategies on MBNAT with ResNet26.}
\vspace{-1em}
\label{fig:mbn-mixing}
\end{figure}

The results in \autoref{fig:mbn-mixing} show that ``MBNAT (logits)" achieves the best robustness-accuracy trade-off curve among all compared interpolation strategies, so we use it as our default interpolation strategy for MBNAT. 

\section{More Experimental Results}
\label{sec:appx-results}

\subsection{Histograms of Decision Margin, Boundary Thickness, and Model Smoothness}
\label{sec:appx-hist}
\begin{figure*}[th] 
	\centering
	\setlength{\tabcolsep}{1pt}
	\begin{tabular}{ccc}
    	\includegraphics[width=0.329\linewidth]{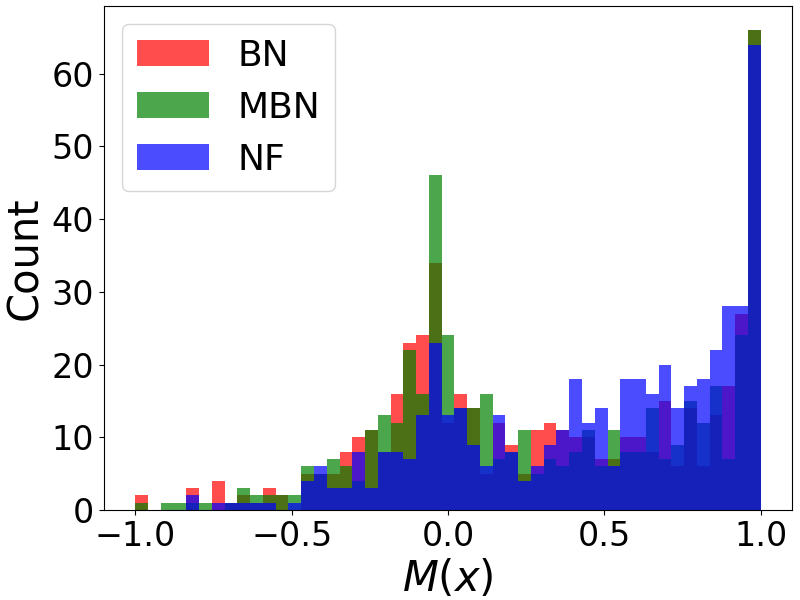} &
    	\includegraphics[width=0.329\linewidth]{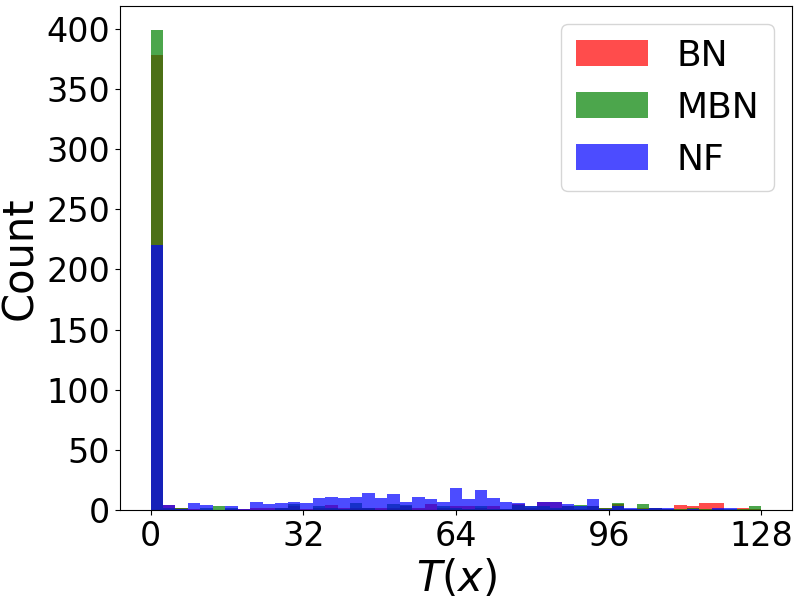} &
    	\includegraphics[width=0.329\linewidth]{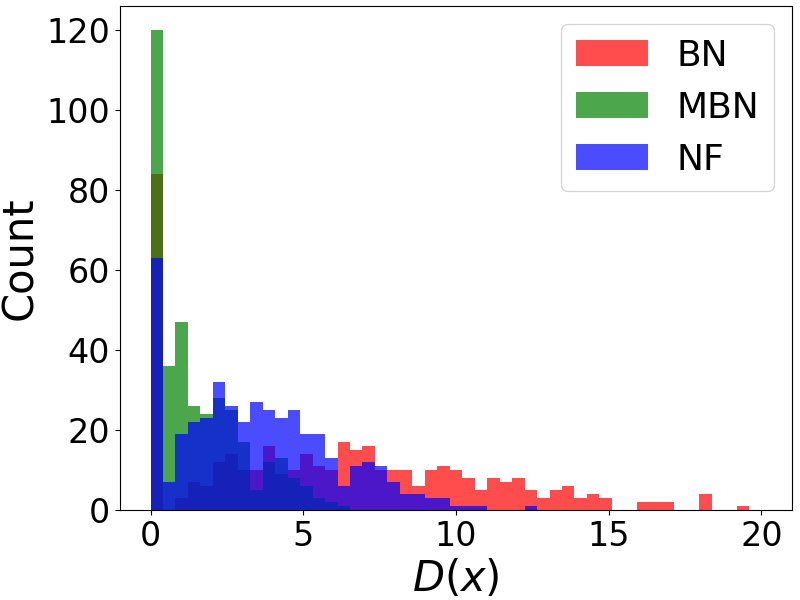}
    	\\
    	(a) Decision margin & (b) Boundary thickness & (c) Model smoothness
		\\
	\end{tabular}
	\vspace{-1em}
	\caption{Histograms of decision margin, boundary thickness, and model smoothness of adversarially trained ResNet26 with different normalization strategies.
	All metrics are measured on the 500 validation images from the first 10 classes of ImageNet.
	(a) $M(\vx)$ as a metric for decision margin: the larger the better.
	(b) $T(\vx)$ as a metric for boundaries robustness: the larger the better.
	(c) $D(\vx)$ as as metric for model smoothness: the smaller the better.
	}
	\vspace{-1em}
	\label{fig:metrics}
\end{figure*}
We have numerically compared the average decision margins, boundary thickness and model smoothness of SAT, MBNAT and NoFrost in \autoref{tab:metrics} (Section \ref{sec:visualization}).
Here in \autoref{fig:metrics}, we visualize the distributions of these metrics on different models using histograms. 
\autoref{fig:metrics} is simply another way to show the results in \autoref{tab:metrics}, but gives more detailed information through histogram visualization.  

\subsection{Normalizer-free Networks with Standard Training is Not Robust to Adversarial Attacks}
\begin{table}[ht]
\centering
\caption{Normalizer-free networks trained using standard training (only on clean images) is not robust against adversarial attacks. All methods are trained on ImageNet with (NF-)ResNet50.}
\resizebox{0.8\linewidth}{!}{
\begin{tabular}{ c|c|c|c }
\toprule
\makecell{\textbf{Method}} & \makecell{\textbf{Clean}} & \makecell{\textbf{PGD}} & \makecell{\textbf{APGD-CE}}  \\
\midrule
ST w/ ResNet50 & \textbf{76.06} & 0.04 & 0.00 \\
ST w/ NF-ResNet50 & \underline{75.02} & 0.00 & 0.00  \\
NoFrost & 74.06 & \textbf{22.45} & \textbf{11.96} \\
\bottomrule
\end{tabular}
}
\label{tab:nfst}
\end{table}
In this section, we show that normalizer-free networks does not naturally have satisfactory adversarial robustness after standard training (i.e., training only on clean images). 
Specifically, we compare the results of standard training on ResNet50 (denoted as ST w/ ResNet50), standard training on NF-ResNet50 (denoted as ST w/ NF-ResNet50), and adversarial training on NF-ResNet50 (NoFrost) in \autoref{tab:nfst}.
As we can see, standard training on both ResNet50 and NF-ResNet50 have almost zero adversarial robustness.
This shows that the robustness of NoFrost is not simply the result of more robust network structure, but the combination of the AT algorithm and the AT-friendly normalizer-free network structure. 

\subsection{Robustness under Different Perturbation Magnitudes}
\begin{figure}[ht]
\centering
\includegraphics[width=0.95\linewidth]{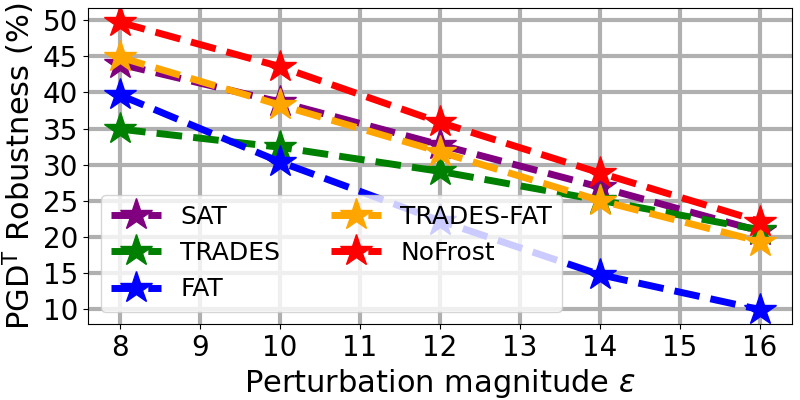}
\vspace{-1em}
\caption{Adversarial robustness under different perturbation magnitudes. All methods are trained on ImageNet with ResNet26.}
\vspace{-1em}
\label{fig:abla-eps}
\end{figure}

In the main text, we evaluated adversarial robustness using adversarial attacks with perturbation magnitude $\eps=8$. 
In this section, we evaluate model robustness under different adversarial perturbation magnitudes.
Specifically, we compare the robustness of the models in \autoref{tab:resnet26} (which are trained with $\eps=8$) on targeted PGD (denoted as $\text{PGD}^{\text{T}}$) attack with $\eps$ ranging from $8$ to $16$.
As shown in \autoref{fig:abla-eps}, the advantage of NoFrost holds on multiple different perturbation magnitudes.

\subsection{Adversarial Training Results with  Small Perturbation Magnitudes}
\label{sec:nfat-results-simple}

In the main text, we set perturbation magnitude $\eps=8$ in both adversarial training and evaluation. 
Some previous works, such as FastAT \cite{wong2019fast} and FreeAT \cite{shafahi2019adversarial} conducted adversarial training on ImageNet using smaller perturbation magnitudes such as $\eps=2,4$, and also evaluation adversarial robustness using the same $\epsilon$ values.
In this section, we compare NoFrost with FastAT and FreeAT using the small $\epsilon$ setting.
We use PGD attack with $10$ and $50$ steps (denoted as PGD-10 and PGD-50 respectively) to evaluate adversarial robustness, following \cite{shafahi2019adversarial}.
The results are shown in \autoref{tab:resnet50-eps2}.
NoFrost largely outperforms FastAT and FreeAT under the small $\eps$ setting. 

\begin{table}[ht]
\vspace{-0.5em}
\centering
\caption{Adversarial robustness of ResNet50 under perturbation magnitude $\eps=2,4$. Classification accuracy on clean images and under different adversarial attacks are reported. The best and second to the best numbers are shown in bold and underlined, respectively.}
\resizebox{0.8\linewidth}{!}{
\begin{tabular}{ c|c|c|c|c}
\toprule
$\epsilon$ & \makecell{\textbf{Method}} & \makecell{\textbf{Clean}} & \makecell{\textbf{PGD-10}} & \makecell{\textbf{PGD-50}}  \\
\midrule
\multirow{3}{*}{2} & FastAT & 60.90 & 44.27 & 44.20 \\
& FreeAT & 64.45 & 43.52 & 43.39 \\
& NoFrost & \textbf{69.87} & \textbf{48.60} &  \textbf{48.23} \\
\midrule
\multirow{3}{*}{4} & FastAT & 55.45 & 32.10 & 31.67 \\
& FreeAT & 60.21 & 32.77 & 31.88 \\
& NoFrost & \textbf{66.05} & \textbf{36.14} & \textbf{36.05} \\
\bottomrule
\end{tabular}
}
\label{tab:resnet50-eps2}
\end{table}


\end{document}

%% file: math_commands.tex
\usepackage{amsmath,amsfonts,bm}

\def\eqref#1{equation~\ref{#1}}

\def\1{\bm{1}}

\def\eps{{\epsilon}}

\def\vtheta{{\bm{\theta}}}

\def\vf{{\bm{f}}}

\def\vp{{\bm{p}}}

\def\vx{{\bm{x}}}

\def\vz{{\bm{z}}}

\def\mW{{\bm{W}}}

\DeclareMathAlphabet{\mathsfit}{\encodingdefault}{\sfdefault}{m}{sl}
\SetMathAlphabet{\mathsfit}{bold}{\encodingdefault}{\sfdefault}{bx}{n}

